\documentclass[letterpaper]{article} 
\usepackage{lmodern}

\usepackage{aaai2026}  
\usepackage{times}  
\usepackage{helvet}  
\usepackage{courier}  
\usepackage[hyphens]{url}  
\usepackage{graphicx} 
\usepackage{natbib}  
\usepackage{caption} 
\usepackage{algorithm}
\usepackage{algorithmic}
\usepackage{amsfonts}
\usepackage{enumitem}
\usepackage{MnSymbol} 
\usepackage{amsmath}
\usepackage{bm}
\usepackage{dsfont}
\usepackage{multirow}
\usepackage{subfigure}
\usepackage{pifont}
\usepackage{xcolor}
\usepackage{booktabs}
\usepackage{xspace}

\newcommand{\m}{TrustEnergy\xspace}

\definecolor{darkgreen}{rgb}{0.0, 0.5, 0.0}
\newcommand\true{\textcolor{darkgreen}{\ding{51}}}
\definecolor{darkred}{rgb}{0.76, 0.13, 0.28}
\newcommand\false{\textcolor{darkred}{\ding{55}}}

\usepackage{newfloat}
\usepackage{listings}
\DeclareCaptionStyle{ruled}{labelfont=normalfont,labelsep=colon,strut=off} 
\floatstyle{ruled}
\newfloat{listing}{tb}{lst}{}
\floatname{listing}{Listing}
\title{\m: A Unified Framework for Accurate and Reliable \\User-level Energy Usage Prediction}

\author{
Dahai Yu\textsuperscript{\rm 1},
Rongchao Xu\textsuperscript{\rm 1},
Dingyi Zhuang\textsuperscript{\rm 2},
Yuheng Bu\textsuperscript{\rm 3},
Shenhao Wang\textsuperscript{\rm 4},
Guang Wang\textsuperscript{\rm 1}\thanks{Prof. Guang Wang is the corresponding author.}
}
\affiliations{
\textsuperscript{\rm 1}Florida State University, Tallahassee, Florida\\
\textsuperscript{\rm 2}Massachusetts Institute of Technology, Cambridge, Massachusetts\\
\textsuperscript{\rm 3}University of California, Santa Barbara, Santa Barbara, California\\
\textsuperscript{\rm 4}University of Florida, Gainesville, Florida\\
dahai.yu@fsu.edu, rxu@fsu.edu, dingyi@mit.edu, buyuheng@ucsb.edu, shenhaowang@ufl.edu, guang@cs.fsu.edu \\
}

\begin{document}

\maketitle

\begin{abstract}
Energy usage prediction is important for various real-world applications, including grid management, infrastructure planning, and disaster response.
Although a plethora of deep learning approaches have been proposed to perform this task, most of them either overlook the essential spatial correlations across households or fail to scale to individualized prediction, making them less effective for accurate fine-grained user-level prediction.
In addition, due to the dynamic and uncertain nature of energy usage caused by various factors such as extreme weather events, quantifying uncertainty for reliable prediction is also significant, but it has not been fully explored in existing work.
In this paper, we propose a unified framework called \m for accurate and reliable user-level energy usage prediction. There are two key technical components in \m, (i) a Hierarchical Spatiotemporal Representation module to efficiently capture both macro and micro energy usage patterns with a novel memory-augmented spatiotemporal graph neural network, and (ii) an innovative Sequential Conformalized Quantile Regression module to dynamically adjust uncertainty bounds to ensure valid prediction intervals over time, without making strong assumptions about the underlying data distribution.
We implement and evaluate our \m framework by working with an electricity provider in Florida, and the results show our \m can achieve a 5.4\% increase in prediction accuracy and 5.7\% improvement in uncertainty quantification compared to state-of-the-art baselines. 
\end{abstract}

\begin{links}
\link{Code}{https://github.com/UFOdestiny/TrustEnergy}
\end{links}

\section{Introduction}\label{introduction}
Energy usage prediction has attracted significant interest from both industry and academia due to its pivotal role in supporting a wide range of practical applications for societal impacts. For example, accurate energy usage prediction is essential for efficient grid management~\cite{JHA2021107479}, outage prevention, and energy distribution optimization. 
It also contributes to emergency preparedness~\cite{zhong2010challenges} by helping stakeholders respond to potential surges in electricity usage during extreme weather events or crises, ensuring the stability and resilience of the power grid.

Although many approaches have been proposed for energy usage prediction by leveraging sophisticated deep learning architectures~\cite{luo2024stacking, xu2024framework, balachander2024innovative, buratto2024seq2seq, aguilar2021short, khan2022efficient, abumohsen2023electrical, yazici2022deep, bashir2022short}, 
most of them primarily focus on time series prediction while overlooking the intricate spatial correlations between different households, which can provide useful information for accurate prediction, e.g., households in close proximity may share similar usage patterns.
Also, most existing works~\cite{stgcn,tu2024powerpm} predict energy usage at a coarse-grained level (e.g., in a city or region level) and fail to capture fine-grained individual household usage patterns for user-level prediction, which potentially reduces their applicability. 
Furthermore, quantifying prediction uncertainty is also important, as it provides prediction intervals and confidence levels, enabling stakeholders to mitigate the risks associated with over- or under-predictions. 
Although there are some methods for spatiotemporal uncertainty quantification\cite{stzinb-gnn, stuanet, wen2023diffstg, yu2025uqgnn}, they fail to capture intrinsic time-varying uncertainty and potential distribution shifts to ensure valid prediction intervals over time.

In this project, by collaborating with a utility provider to improve its energy supply services, we are fortunate to have access to a long-term fine-grained household-level electricity usage dataset, which provides us with a good opportunity to explore user-level energy usage prediction. However, we found that there are two practical challenges. 
(i) Modeling energy usage at the individual user level is computationally costly when there are hundreds of thousands of users. For example, there are over 60,000 users in our dataset, each exhibiting unique temporal usage patterns and spatial correlations with others. Modeling spatiotemporal usage of each user explicitly would not only result in an intractable computational burden but also pose serious challenges in terms of data sparsity and overfitting. 
(ii) Spatiotemporal usage patterns are intricately dynamic and time-varying, which are impacted by different factors like seasonal trends, weather conditions, and personal habits. This dynamics potentially causes data distribution shifts and high prediction uncertainty. Effectively modeling such uncertainty remains nontrivial but is essential for delivering reliable predictions.

To address the above two challenges and advance existing works, we propose a unified framework named \m for accurate and reliable energy usage prediction.
\m~ consists of two key novel components: (i) a Hierarchical Spatiotemporal Representation module (HSTR), which is capable of efficiently capturing both micro user-level and macro region-level patterns based on a new memory-augmented spatiotemporal graph neural network (MASTGNN), and (ii) a Sequential Conformalized Quantile Regression module (SCQR), which is a distribution-agnostic uncertainty quantification method that dynamically adjusts uncertainty bounds to provide valid prediction intervals over time for reliable prediction without making strong assumptions about the underlying data distributions.
The key contributions of this paper are as follows:
\begin{itemize}
    \item Conceptually, we frame household-level energy usage prediction as a spatiotemporal prediction problem rather than a traditional time-series prediction task. We aim to improve not only the accuracy but also the reliability of energy usage predictions for real-world societal impact by incorporating both macro-level regional patterns and micro-level user behaviors.
    
    \item Technically, we design a unified framework called \m for accurate and reliable energy usage prediction, which includes two key novel modules: (i) a Hierarchical Spatiotemporal Representation Learning module with an innovative MASTGNN to efficiently and effectively capture comprehensive energy usage patterns from both micro level and macro level, and (ii) a distribution-agnostic uncertainty quantification module called SCQR to dynamically adjust uncertainty bounds to ensure valid prediction intervals over time.

    \item Empirically, we extensively evaluate our \m by collaborating with one of the largest municipal utilities in Florida. We compare \m with 13 state-of-the-art baselines across six metrics, and the experimental results show the superiority of \m, e.g., it increases prediction accuracy by 5.4\% and uncertainty quantification by 5.7\% over the best baseline on the real-world data from our collaborator. We also verify the generalizability of our \m using two other datasets from New York and California.
\end{itemize}


\section{Problem Formulation}\label{preliminary}
In this section, we formally formulate the energy usage prediction problem as a spatiotemporal graph learning problem since a graph can capture comprehensive dependencies between different entities (e.g., users) in the energy system. 

\subsubsection{Hierarchical Graphs Construction} \label{sec:network}
Formally, a graph can be defined as $\mathcal{G}=\left( \mathcal{V}, \mathcal{E}, \mathcal{A} \right)$, where $\mathcal{V}=\left( v_1, v_2, ..., v_N\right)$ is a set of 
$\lvert \mathcal{V} \rvert:=N$ nodes, $\mathcal{E}=\left( e_1, e_2, ...\right)$ is a set of edges, and $\mathcal{A}$ is an adjacency matrix. 
The adjacency matrix derived from this graph can be denoted as $\mathcal{A}\in \mathbb{R}^{N\times N}$, where $\mathcal{A}_{v_i,v_j}$ denotes the edge weight between node $v_i$ and node $v_j$,
which can be denoted as: 
\begin{equation}
\mathcal{A}_{v_i,v_j}=\left\{\begin{array}{l} \exp(-\frac{d_{ij}^2}{\sigma^2}), i \neq j~\text{and}~\exp(-\frac{d_{ij}^2}{\sigma^2})>=r
\\ 0\quad\quad\quad\quad,\text{otherwise}.
\end{array}\right.,
\end{equation}
where $d_{ij}$ represents the distance of the centroids between node $v_i$ and node $v_j$. $\sigma^2$ and $r$ are thresholds to control the distribution and sparsity of the matrix, where a large $r$ value can be used to accelerate the model training process. 

Based on our data analysis, a user's energy usage can be predicted not only from their historical data but also by leveraging the usage patterns of nearby users or those with similar usage behaviors.
Therefore, instead of constructing a single graph, we construct two different graphs to comprehensively learn hierarchical patterns from the user aspect and region aspect, which capture energy usage patterns at the micro level and macro level, respectively. 
We first construct a \textbf{micro-level graph} $\mathcal{G}_u=\left( \mathcal{V}_u, \mathcal{E}_u, \mathcal{A}_u \right)$ as defined before to model the correlation between all users, where each user is a node, and the edge is the geographical distance between households.
Similarly, a \textbf{macro-level graph} is defined as $\mathcal{G}_r=\left( \mathcal{V}_r, \mathcal{E}_r, \mathcal{A}_r \right)$, where each node denotes a geospatial region, and each edge means the geographical distance between two regions.

\begin{figure*}[tb]
  \centering
  \includegraphics[width=\linewidth]{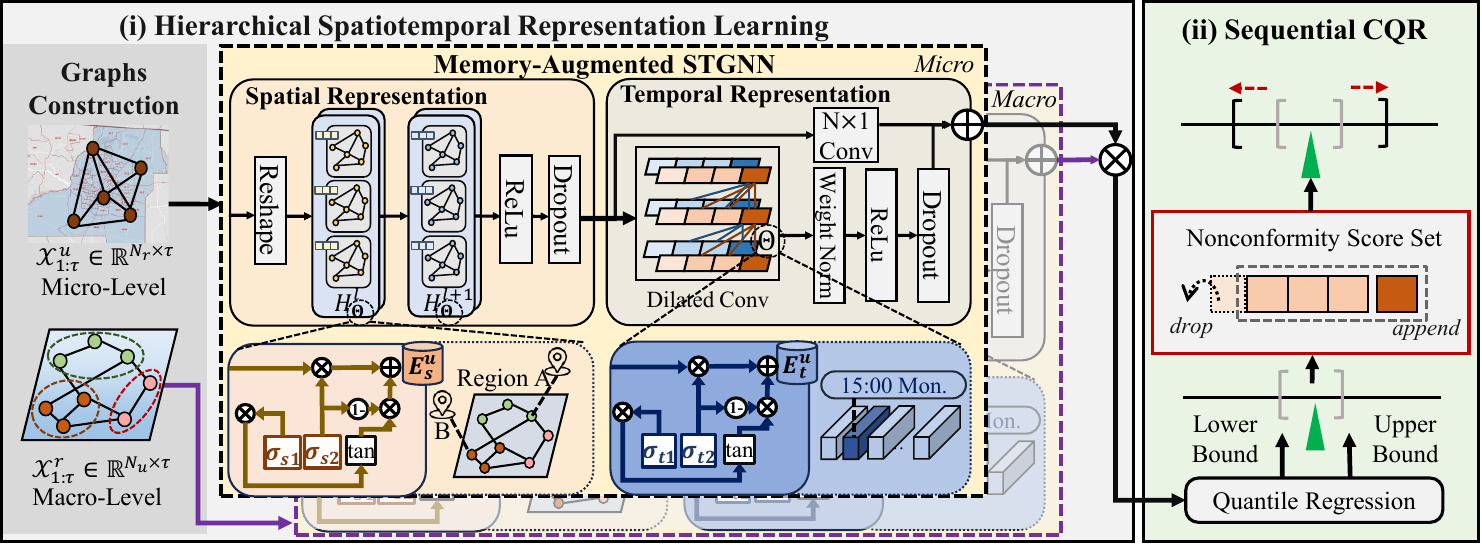}
  \caption{
  An overall framework of \m, which consists of (i) a Hierarchical Spatiotemporal Representation Learning module to efficiently capture both macro-level and micro-level patterns, and (ii) a Sequential Conformalized Quantile Regression module to dynamically adjust uncertainty bounds to ensure valid prediction intervals over time.
  }
  \label{framework}
\end{figure*}

\subsection{Energy Usage Prediction}\label{Electricity Load Prediction}
Given the above-constructed hierarchical graphs, we use $\mathcal{X}^u$ and $\mathcal{X}^r$ to denote the micro-level and macro-level graph signals, respectively. 
Suppose we use $x^i_t\in \mathbb{R}$ to denote the value of node $i$ at time $t$ and $X_{t}=(x^{1}_{t},x^{2}_{t},...,x^{N}_{t}) \in \mathbb{R}^{N}$ denotes the values of all the $N$ nodes at time $t$. Then, historical data of $\tau$ time steps can be defined $\mathcal{X}^u_{1:\tau}=(X^u_{1}, X^u_{2},..., X^u_{\tau}) \in \mathbb{R}^{N^u\times \tau}$ for micro-level graph, and $\mathcal{X}^r_{1:\tau}=(X^r_{1}, X^r_{2},..., X^r_{\tau}) \in \mathbb{R}^{N^r\times \tau}$ for macro-level graph. Conventional energy usage prediction aims to learn a mapping function $f$ from historical data to predict values at the following $T$ time steps, which can be expressed as: 
\begin{equation}
[\mathcal{X}^u_{1:\tau},\mathcal{X}^r_{1:\tau}]\stackrel{f}{\longrightarrow}\hat{\mathcal{X}^u}_{\tau+1:\tau+T},\label{eq.2}
\end{equation}
where $\hat{\mathcal{X}^u}_{\tau+1:\tau+T} \in \mathbb{R}^{N\times T}$. 

Instead of solely considering prediction accuracy by outputting a single expected value, we also aim to enhance prediction reliability by providing a prediction interval distribution. 
A common approach to achieve this is the quantile regression, which outputs different quantiles (e.g., 90th percentile) of the predicted variable under the miscoverage rate of $\alpha$, where we define $q_{\alpha_{\text{lo}}}$, $q_{\alpha_{\text{up}}}$, and $q_{\alpha_{\text{mi}}}$ as lower quantile, upper quantile, and median, respectively.
The uncertainty-aware (i.e., probabilistic) prediction can be expressed as:
\begin{equation}
[\mathcal{X}^u_{1:\tau},\mathcal{X}^r_{1:\tau}]\stackrel{\mathcal{F}}{\longrightarrow}[\hat{L}_{\tau+1:\tau+T},\hat{U}_{\tau+1:\tau+T},\hat{M}_{\tau+1:\tau+T}],
\end{equation}
where $\hat{L}_{\tau+1:\tau+T}=\hat{q}_{\alpha_{\text{lo}}}(\mathcal{X}^u_{1:\tau})\in \mathbb{R}^{N^u\times T}$ is the lower quantile prediction, $ \hat{U}_{\tau+1:\tau+T}=\hat{q}_{\alpha_{\text{up}}}(\mathcal{X}^u_{1:\tau})\in \mathbb{R}^{N^u\times T}$ is the upper quantile prediction, and $\hat{M}_{\tau+1:\tau+T}=\hat{q}_{\alpha_{\text{mi}}}(\mathcal{X}^u_{1:\tau})\in \mathbb{R}^{N^u\times T}$ is the median prediction, equivalent to $\hat{\mathcal{X}^u}_{\tau+1:\tau+T}$ in Eq. \ref{eq.2}.

\section{Methodology}\label{methodology}
In this part, we introduce the detailed design of our \m. The overview framework is shown in Figure~\ref{framework}.

\subsection{Hierarchical Spatiotemporal Representation}

In this work, we focus on user-level energy usage prediction, which is computationally expensive for existing spatiotemporal graph learning frameworks~\cite{tu2024powerpm, uqunifi, wen2023diffstg, GraphWavenet, dong2024heterogeneity} to model fine-grained spatial and temporal patterns due to large-scale users and huge timesteps. For example, in our problem, there are $T=17,520$ 30-minute time slots in a year and over $N=60,000$ households, so the parameter space is at a billion scale.
Directly optimizing parameters in such a large parameter space is computationally expensive.
To address this issue, inspired by  HimNet~\cite{dong2024heterogeneity}, we propose a memory-augmented spatiotemporal graph neural network (MASTGNN) to efficiently and effectively capture hierarchical spatiotemporal patterns.

MASTGNN includes a memory-augmented mechanism based on a well-designed shared parameter pool to update internal states, which significantly reduces model computational complexity. 
Specifically, the memory denotes a shared parameter pool that stores a fixed number of trainable parameter blocks. At each update step, task-specific parameters are dynamically constructed by retrieving and aggregating relevant blocks from this pool. 
This memory-augmented mechanism can be adapted to any STGNN framework. Without loss of generality, we employ an effective backbone combining Diffusion Graph Convolutional Networks \cite{dcrnn} and Temporal Convolutional Networks \cite{stzinb-gnn} to model the spatial and temporal patterns, and our memory-augmented mechanism will be utilized to efficiently update model parameters.

\subsubsection{Parameter Pool Construction}
We construct parameter pools from both spatial and temporal dimensions to model user behaviors. 
We build a spatial embedding matrix $E^u_s \in \mathbb{R}^{N^u\times d_s}$, where $N^u$ denotes the total number of users in the dataset, $d_s$ is the dimension of the embedding vector assigned to each location. Similarly, the spatial embedding for the macro-level graph is $E^r_s \in \mathbb{R}^{N^r\times d_s}$, where $N^r$ denotes the total number of regions. 
From the temporal dimension, we found a significant energy usage behavior shift for different hours, days, and months. Therefore, we construct a time-of-day embedding $D_{\text{tod}} \in \mathbb{R}^{N_\text{tod}\times d_\text{tod}}$, a day-of-week embedding $D_{\text{dow}} \in \mathbb{R}^{N_\text{dow}\times d_\text{dow}}$, and a month-of-year embedding $D_{\text{moy}} \in \mathbb{R}^{N_\text{moy}\times d_\text{moy}}$, where $d_\text{tod}$, $d_\text{dow}$ and $d_\text{moy}$ denotes the number of time steps per day, the number of days in a week and the number of months in a year, respectively. 
Typically, $d_\text{tod}$ is dependent on the original time interval, $d_\text{dow}=7$, and $d_\text{moy}=12$. 
The last sample’s time information (tod, dow, moy) of each batch is used as the index to extract the corresponding embedding $E_{\text{tod}} \in \mathbb{R}^{B\times d_\text{tod}}$, $E_{\text{dow}} \in \mathbb{R}^{B\times d_\text{dow}}$, and $E_{\text{moy}} \in \mathbb{R}^{B\times d_\text{moy}}$.
The three embeddings are concatenated to obtain the overall temporal embedding matrix $E^u_t=E^r_t=E_{\text{tod}}||E_{\text{dow}}||E_{\text{moy}} \in \mathbb{R}^{B\times d_t}$, where $B$ is the batch size and $d_t=d_\text{tod}+d_\text{dow}+d_\text{moy}$.

For the spatial dimension, we define a parameter pool $P_s \in \mathbb{R}^{d_s \times M}$ and generate spatial parameters using the spatial embedding $E_s \in \mathbb{R}^{N \times d_s}$:
\begin{equation}
    \theta_s = E_s \cdot P_s,
\end{equation}
where $\theta_s \in \mathbb{R}^{N \times M}$.
Similarly, the temporal parameter pool is defined as $P_t \in \mathbb{R}^{d_t \times M}$, containing $d_t$ candidate vectors. Given the temporal embedding $E_t \in \mathbb{R}^{B \times d_t}$ (from $B$ samples), the temporal parameters are generated by:
\begin{equation}
    \theta_t = E_t \cdot P_t,
\end{equation}
where $\theta_t \in \mathbb{R}^{B \times M}$.

Furthermore, to efficiently generate parameters for large-scale user sets, we partition the full parameter sets $P_s$ and $P_t$ into multiple smaller blocks. Take $P_s$ as an example:
\begin{equation}
P_s = [P_{s_1}, P_{s_2}, \ldots, P_{s_I}], \quad P_{s_i} \in \mathbb{R}^{d_s \times m_{s_i}}, \quad \sum_{i=1}^I m_{s_i} = M.
\end{equation}
Each block $P_{s_i}$ is represented by a centroid vector $\mu_i \in \mathbb{R}^k$, which summarizes the block's parameter distribution.
Given a query $E_s$, we first perform a coarse selection step by measuring the similarity between $E_s$ and each centroid $\mu_i$. The index of the most similar block is obtained by:
\begin{equation}
i = \arg\max_i \text{sim}(E_s, \mu_i),
\end{equation}
where $\text{sim}(\cdot, \cdot)$ denotes a similarity metric such as cosine similarity or dot product.
Finally, the parameter $\theta_s$ is generated by multiplying the query vector $E_t$ with the selected block $P_{s_i}$:
\begin{equation}
    \theta_s = E_s \cdot P_{s_i}.
\end{equation}

By reusing shared parameter pools and querying them adaptively via embeddings, \m effectively reduces the computational complexity from $\mathcal{O}(TN)$ to $\mathcal{O}(k)$, where $k=B+N$ is the size of the parameter pool. This enables efficient and adaptive learning across highly heterogeneous user-level energy usage patterns.

\subsubsection{Parameter Updating Mechanism}
Our method builds upon a memory-augmented recurrent unit as the fundamental component, where both spatial and temporal dynamics are captured via structured gating and message passing.
The fundamental building block is formulated by:
\begin{equation}
\begin{aligned}
&r_t=\sigma \left(\Theta_r\star_G [X_t,H_{t-1}] +b_r \right) \\
&u_t=\sigma \left(\Theta_u\star_G [X_t,H_{t-1}] +b_u  \right)\\
&c_t=\text{tanh} \left(\Theta_c\star_G [X_t,(r_t \odot H_{t-1})]+b_c  \right)\\
&H_t=u_t\odot H_{t-1}+(1-u_t)\odot c_t   \\
\end{aligned}
\end{equation}
where $X_t \in \mathbb{R}^{N}$ and $H_t \in \mathbb{R}^{N \times h}$ denote the input and output at timestep $t$. $h$ is the hidden layer size. $r_t$ and $u_t$ are the reset and update gates. $\Theta_r$, $\Theta_u$, $\Theta_c$ are parameters for the corresponding filters in the graph convolution operation $\star_{\mathcal{G}}$, which is defined as:
\begin{equation}
Z = U \star_{\mathcal{G}} = \sum_{k=0}^{K} \tilde{A}^k U W_k,
\end{equation}
where $U \in \mathbb{R}^{N \times C}$ is the input, $Z \in \mathbb{R}^{N \times h}$ is the output, and $C$ denotes the input channels. $\tilde{A}$ represents the topology of graph $\mathcal{G}$, and $W \in \mathbb{R}^{K \times C \times h}$ is the kernel parameter $\Theta$.
We use temporal embeddings $E_t \in \mathbb{R}^{B \times d_t}$ and spatial embeddings $E_s \in \mathbb{R}^{N \times d_s}$ to dynamically generate context-dependent meta-parameters via low-rank linear attention:
\begin{equation}
W_t = E_t \cdot P_t, \quad W_s = E_s \cdot P_s,
\end{equation}
where $P_t \in \mathbb{R}^{d_t \times K \times C \times h}$ and $P_s \in \mathbb{R}^{d_s \times K \times C \times h}$ are meta-parameter pools for the temporal and spatial dimensions, respectively. $W_t \in \mathbb{R}^{B \times K \times C \times h}$ and $W_s \in \mathbb{R}^{N \times K \times C \times h}$ are the corresponding meta-parameters.
Given input sequences $X \in \mathbb{R}^{B \times T \times N}$, the model computes hidden states $H \in \mathbb{R}^{B \times T \times N \times h}$ using the generated $W_t$ and $W_s$. These hidden states are directly passed to a lightweight prediction head (e.g., linear layer or MLP) to obtain the final prediction output.

\subsubsection{Loss Function}
Since our target is to output the lower quantile, upper quantile, and median of the prediction, 
we design a hybrid loss function by combining both the pinball loss $\rho_\alpha(Y,\hat{X})$ for quantile regression and mean absolute error (MAE) loss for median prediction, which is denoted as:
\begin{equation}
    \begin{aligned}
\mathcal{L}(Y,\hat{X}) =&\rho_{\alpha_{\text{lo}}}(Y,\hat{X})+\rho_{\alpha_{\text{up}}}(Y,\hat{X})\\
&+\text{MAE}(Y,\hat{q}_{\alpha_{\text{mi}}}(X)),
    \end{aligned}
\end{equation}
where $\rho_\alpha(Y,\hat{X})$ is calculated: 
\begin{equation}
    \begin{aligned}
\rho_\alpha(Y,\hat{X})=\left\{\begin{array}{l} \alpha(Y-\hat{X})  \quad\quad\quad\text{if}~Y-\hat{X}>0
\\ (1-\alpha)(Y-\hat{X}) \quad\text{otherwise}
\end{array}\right. .
    \end{aligned}
\end{equation}

\subsection{Sequential Conformalized Quantile Regression}

Prediction intervals from traditional quantile regression may be poorly calibrated in energy usage prediction scenarios due to distribution shifts caused by factors such as seasonal changes, weather fluctuations, and behavioral changes, which can lead to miscoverage and the failure to provide reliable uncertainty estimates.
To address this challenge, we propose Sequential Conformalized Quantile Regression (SCQR), a distribution-free uncertainty quantification method that ensures valid prediction intervals with guaranteed coverage.
SCQR dynamically adjusts uncertainty bounds over time, enhancing quantile regression and maintaining a valid coverage rate without strong assumptions about the underlying data distribution.

Given a dataset from previous $n$ timesteps $\{(X_i, Y_i)\}^n_{i=1}$, we first split it into a training set $\mathcal{D}_\text{tr}$, a calibration set $\mathcal{D}_\text{ca}$, and a test set $\mathcal{D}_\text{te}$.
We then obtain the nonconformity scores set $\mathcal{E}=\{\hat{\epsilon}_1,\hat{\epsilon}_2,...\hat{\epsilon}_{t}\}$, where $t=|\mathcal{D}_{ca}|$ and $\hat\epsilon_i$ is computed by nonconformity measure $S$:
\begin{equation}\small
\hat\epsilon_i=S\left\{(X_{i}, Y_{i})\right\}=\text{max}\left\{\hat{q}_{\alpha_{\text{lo}}}(X_i)-Y_i, Y_i-\hat{q}_{\alpha_{\text{up}}}(X_i) \right\},\label{eq.4}
\end{equation}
where $ i\in \mathcal{D}_\text{ca}$. To conformalize the prediction interval, we compute $Q_{1-\alpha}(\mathcal{E},\mathcal{D}_\text{ca})$ on the set $\mathcal{E}$:
\begin{equation}\small
Q_{1-\alpha}(\mathcal{E},\mathcal{D}_\text{ca})=\text{quantile}\left( (1-\alpha)(1+1/|\mathcal{D}_\text{ca}|)\right)~\text{of set }\mathcal{E}.\label{eq.5}
\end{equation}
Finally, given new input data $X_{new}$, we set $\text{low}=\hat{q}_{\alpha_{\text{lo}}}(X_{new})$ and $\text{high}=\hat{q}_{\alpha_{\text{up}}}(X_{new})$. We then construct the prediction interval for $Y_{new}$ as:
\begin{equation}\small
\hat{C}(X_{new}) = [\text{low} - Q_{1-\alpha}(\mathcal{E},\mathcal{D}_\text{ca}),
\text{high} + Q_{1-\alpha}(\mathcal{E},\mathcal{D}_\text{ca})].\label{eq.6}
\end{equation}

\begin{algorithm}[tb]
\caption{Sequential\hspace{-0.017em} Conformalized\hspace{-0.017em} Quantile\hspace{-0.017em} Regression}
\begin{algorithmic}[1]

\REQUIRE Training set $\mathcal{I}_{tr}$, calibration set $\mathcal{I}_{ca}$, test set $\mathcal{I}_{te}$; a quantile regressor $\mathcal{F}$; target error rate $\alpha$; nonconformity measure $S$.

\ENSURE Prediction intervals $\hat{C}(X_{t})$ for all $t \in \mathcal{I}_{te}$.

\STATE Train the quantile regressor to obtain $\left\{\hat{q}_{\alpha_{\text{lo}}}, \hat{q}_{\alpha_{\text{up}}}\right\} \gets \mathcal{F}\left(\left\{(X_i, Y_i)\right\}_{i\in \mathcal{I}_{tr}}\right)$.

\STATE Compute nonconformity scores $\mathcal{E} = \{\hat{\epsilon}_1, \hat{\epsilon}_2, \dots, \hat{\epsilon}_{t}\}$, where $t = |\mathcal{I}_{ca}|$ and $\hat{\epsilon}_i = S((X_i, Y_i))$ for $i \in \mathcal{I}_{ca}$, as defined in Eq.~\ref{eq.4}.

\FOR{each time step $t \in \mathcal{I}_{te}$}
\STATE Compute the quantile $Q_{1-\alpha}(\mathcal{E},\mathcal{D}_\text{ca})$, as in Eq.~\ref{eq.5}.

\STATE Construct the prediction interval $\hat{C}(X_{t})$, as in Eq.~\ref{eq.6}.

\STATE Obtain new nonconform score $\hat{\epsilon}_t {\longleftarrow}S\left((X_t, Y_t)\right)$.

\STATE Update the new set $\mathcal{E}$ by removing the oldest element and adding the newest $\hat{\epsilon}_t$.

\ENDFOR
\end{algorithmic}
\label{SCQR}
\end{algorithm}

To comprehensively capture the underlying temporal dependencies inherent in sequential data, SCQR explicitly incorporates the historical behavior of residuals (i.e., nonconformity scores) when constructing valid prediction intervals. This approach is different from traditional conformal prediction paradigms that assume exchangeability and independence among samples by sequentially adapting the conformal score set to reflect the temporal structure of the data stream.
Formally, consider a test-time scenario where a sequence of test samples $\{(X_{t+1}, Y_{t+1}), (X_{t+2}, Y_{t+2}), \dots\}$ arrives in a streaming or temporally ordered fashion. Let $\mathcal{E} = \{\hat{\epsilon}_1, \hat{\epsilon}_2, \dots, \hat{\epsilon}_t\}$ denote the current set of nonconformity scores derived from a held-out calibration set $\mathcal{D}\text{ca}$. When the first test sample $(X_{t+1}, Y_{t+1})$ is observed, SCQR computes a new nonconformity score $\hat{\epsilon}_{t+1} = S(X_{t+1}, Y_{t+1})$ using a predefined scoring function $S(\cdot)$, typically based on the deviation between the ground truth $Y_{t+1}$ and the model's predictive quantile estimate at $X_{t+1}$.

To maintain a rolling calibration window and preserve the sequential dependency, the earliest score $\hat{\epsilon}_1$ in $\mathcal{E}$ is discarded, and the newly computed score $\hat{\epsilon}_{t+1}$ is appended to the set, yielding an updated nonconformity set:
$\mathcal{E}=\{\hat{\epsilon}_2,\hat{\epsilon}_3,..., \hat{\epsilon}_{t+1}\}$
This dynamic update mechanism ensures that the conformal quantile $Q_{1-\alpha}(\mathcal{E}, \mathcal{D}_\text{ca})$, which determines the prediction interval width for a given coverage level $1-\alpha$, is recalibrated in real time to accommodate evolving patterns in the residual distribution.
Subsequently, the updated quantile estimate is utilized to compute the prediction interval $\hat{C}(X_{t+2})$ for the next incoming test sample $(X_{t+2}, Y_{t+2})$, maintaining validity under the sequential regime. This recursive procedure continues iteratively, such that the conformal set $\mathcal{E}$ is refreshed at each time step to reflect the most recent residual dynamics. The process is illustrated in Algorithm~\ref{SCQR}, ensuring that each prediction interval is adaptively informed by the most temporally relevant uncertainty information, thereby enhancing calibration fidelity in non-stationary or temporally correlated settings.

\section{Evaluation}\label{evaluation}

In this project, we collaborate with an electricity provider in Florida to evaluate our \m framework.
Specifically, we try to answer the following questions:
\begin{itemize}[leftmargin=5.0mm]
\item \textbf{RQ 1}: How does \m perform compared to state-of-the-art models across different metrics?
\item \textbf{RQ 2}: How does \m perform under abnormal scenarios such as extreme weather events?
\item \textbf{RQ 3}: What is the individual contribution of each module to \m's performance?

\item \textbf{RQ 4}: Can \m generalize to other regions?
\end{itemize}

\subsection{Experimental Setup
}\label{datasets}

\subsubsection{Datasets}
We evaluate the performance of our \m using fine-grained 30-minute interval electricity consumption data for 61,672 households across 253 regions in Florida (FL) in 2018. In addition, we also use two other real-world datasets from New York (NY) \cite{nyiso} and California (CA) \cite{caiso} to verify the generalizability of our design.



\subsubsection{Baselines}
We compare \m with 13 state-of-the-art baselines, including 
Spatial-Temporal Graph Convolution Network (STGCN)~\cite{stgcn}, 
Graph WaveNet (GWNET)~\cite{GraphWavenet}, 
(ASTGCN)~\cite{guo2019attention}, 
Adaptive Graph Convolutional Recurrent Network (AGCRN)~\cite{bai2020adaptive}, 
Spectral Temporal GNN (StemGNN)~\cite{cao2020spectral}, 
Dynamic Spatial-Temporal Aware Graph Neural Network (DSTAGNN)~\cite{dstagnn}, 
STZINB~\cite{stzinb-gnn},
PDFormer~\cite{pdformer}, 
DiffSTG~\cite{wen2023diffstg}, 
DeepSTUQ~\cite{uqunifi},
PowerPM~\cite{tu2024powerpm},
Chronos~\cite{chronos},
and Moment~\cite{moment}.

\subsubsection{Evaluation Metrics}

We utilize three metrics, including Mean Absolute Error (MAE), Root Mean Squared Error (RMSE), and Mean Absolute Percentage Error (MAPE), to evaluate the deterministic performance, and three other metrics, including Mean Prediction Interval Width (MPIW) \cite{stzinb-gnn}, Winkler score (WINK)~\cite{han2023probabilistic}, and coverage (COV)~\cite{zhuang2024sauc}, to evaluate the performance of uncertainty quantification.

\subsection{Overall Performance Comparison (RQ 1)}
\begin{table}[tb]\small
\setlength{\tabcolsep}{3.5pt}
\centering 

\begin{tabular}{l|cccccc}
\toprule
\multirow{2}{*}{Models} & \multicolumn{6}{c}{Metrics} \\  \cline{2-7} 
& MAE & RMSE & MAPE  & MPIW & WINK & Cov\\ 
\hline
STGCN & 49.72 & 59.65 & 1.37\% & 99.75 & 132.17 & \false \\
GWNET & 63.52 & 70.36 & 1.58\% & 115.37 & 138.62 & \false \\
ASTGCN & 56.25 & 64.17 & 1.39\% & 105.28 & 130.49 & \false \\
AGCRN & \underline{48.35} & \underline{58.31} & 1.40\% & 96.23 & 120.85 & \true \\
StemGNN & 51.66 & 63.28 & 1.32\% & 94.05 & 113.43 & \true\\
DSTAGNN & 58.71 & 68.79 & 1.47\% & 107.67 & 125.81 & \false \\
STZINB & 55.31 & 68.43 & \textbf{1.28\%} & 96.70 & 120.63 & \false \\
PDFormer & 53.71 & 66.07 & 1.42\% & 97.31 & 117.20 & \false \\
DiffSTG & 53.28 & 67.91 & 1.30\% & 95.83 & 125.70 & \false \\
DeepSTUQ & 55.75 & 68.05 & 1.32\% & 94.25 & 122.10 & \true \\
PowerPM & 50.25 & 60.11 & \underline{1.29\%} & 99.52 & 135.41 & \false \\
Chronos     &48.50& 59.02 &  1.35\% & \underline{90.94} & \underline{108.57} & \true  \\
Moment      &53.20& 64.75 &  1.38\% & 95.42 & 118.94 & \false  \\
\textbf{\m} & \textbf{45.72} & \textbf{55.65} & \textbf{1.28\%} & \textbf{85.72} & \textbf{102.31} & \true \\

\bottomrule
\end{tabular}
\caption{Comparison with 13 state-of-the-art baselines using Florida data on six metrics. The best results are presented in bold, and the second-best results are underlined. \true~ means that the method reaches the target coverage (i.e., coverage $\geq$ 90\%) while \false~ means that it fails to achieve it.
}\label{result}
\end{table}

\begin{figure}[tb]
\centering
\subfigure[Ideal vs. Empirical Cov]{\label{fig_cov1}
\includegraphics[width=0.48\linewidth]{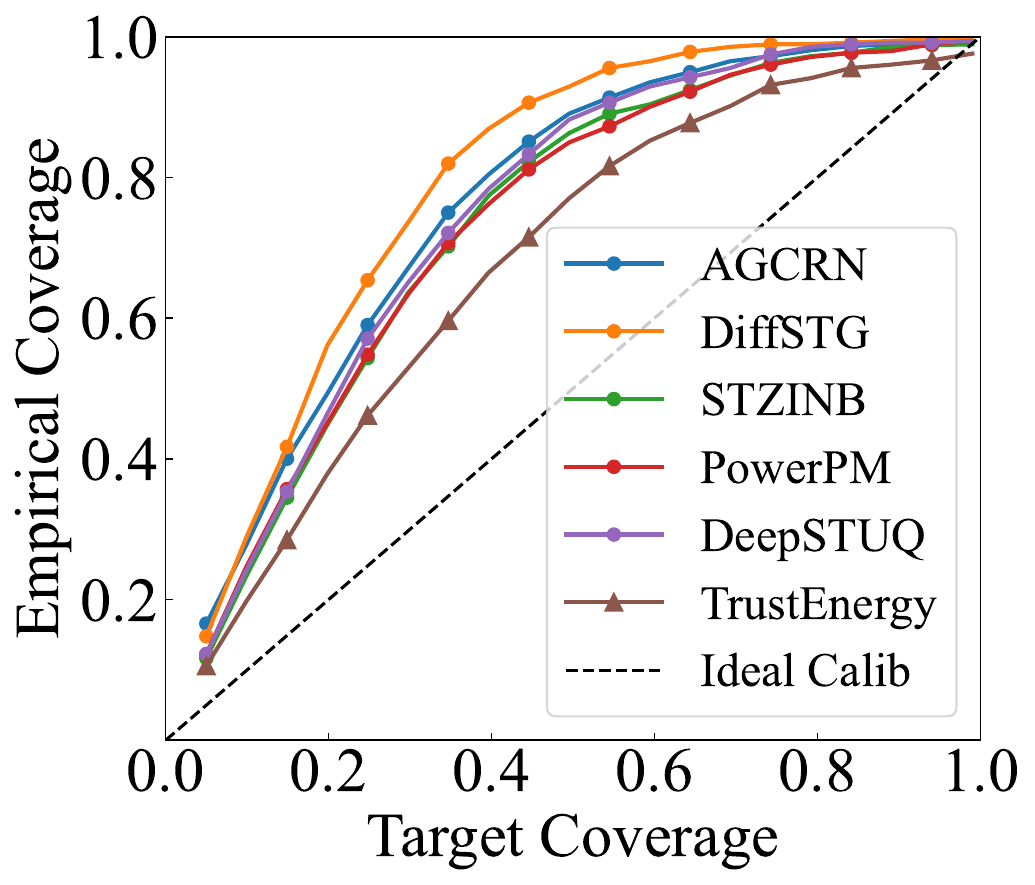}}
\subfigure[Coverage vs. MAE]{\label{fig_cov2}
\includegraphics[width=0.48\linewidth]{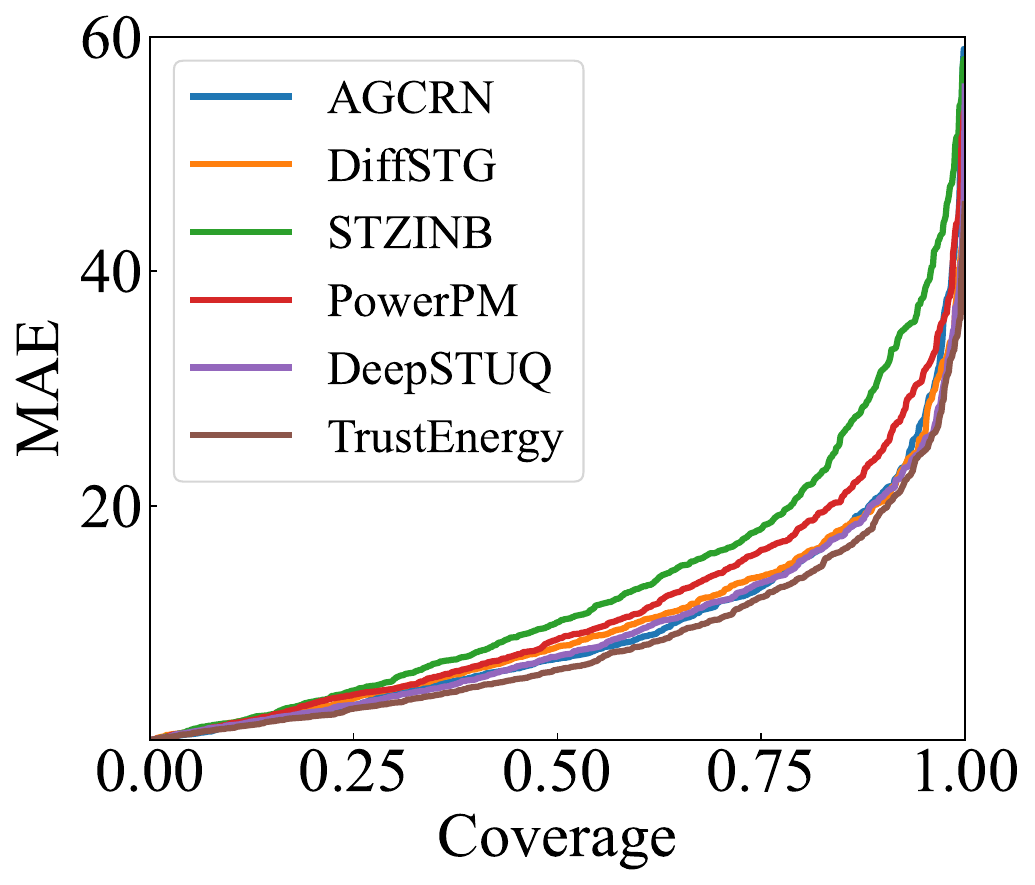}}
\caption{Uncertainty calibration results.}
\label{curve_figure}
\end{figure}

An overall comparison of our \m and other baseline models is presented in Table \ref{result}. We found that our \m consistently achieves the best performance across all metrics for both prediction accuracy and reliability. Specifically, our framework reduces MAE by approximately 5.4\% compared to the best baseline AGCRN.
In addition, \m also demonstrates superior performance on prediction reliability, with a 5.7\% improvement in WINK and reaching the target coverage. Figure~\ref{fig_cov1} compares nominal coverage levels with empirical coverage, which shows \m achieves the most reliable calibration as its curve is closest to the diagonal line. 
Figure~\ref{fig_cov2} also demonstrates the reliability by comparing the coverage and MAE. It is observed that \m achieves a higher coverage with the same MAE level, indicating its better performance for uncertainty quantification.

\subsection{Case Studies for Abnormal Situations (RQ 2)}
One of the most challenging aspects of energy usage prediction lies in handling abnormal or extreme scenarios, such as hurricanes and heat waves, that often lead to sudden and significant deviations in energy usage. 
Accurate and reliable energy usage prediction under these extreme weather events is critical for ensuring grid stability, optimizing energy resource allocation, and enabling timely decision-making for both utilities and emergency response teams.

\begin{figure}[tb]
\centering
\subfigure[User 1 in Hurricane]{
\includegraphics[width=0.48\linewidth]{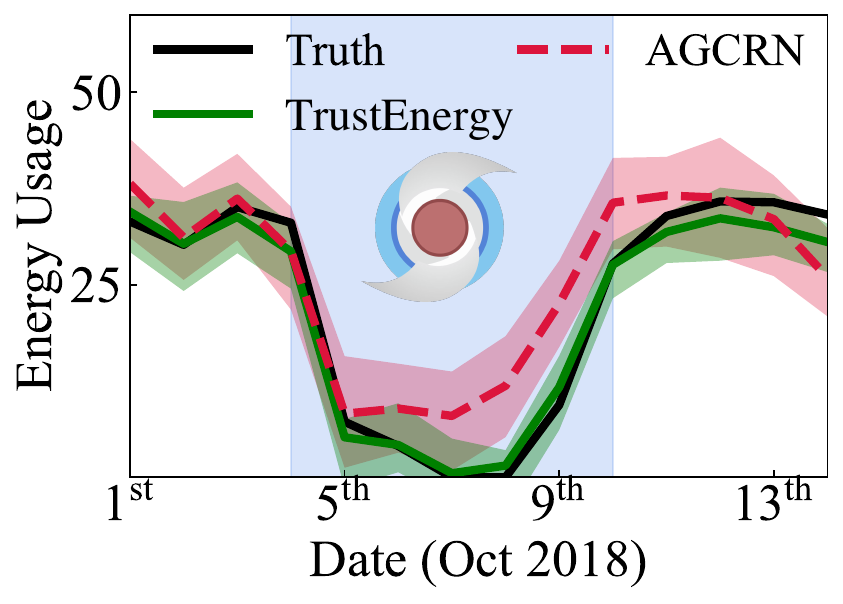}}
\subfigure[User 2 in Hurricane]{
\includegraphics[width=0.48\linewidth]{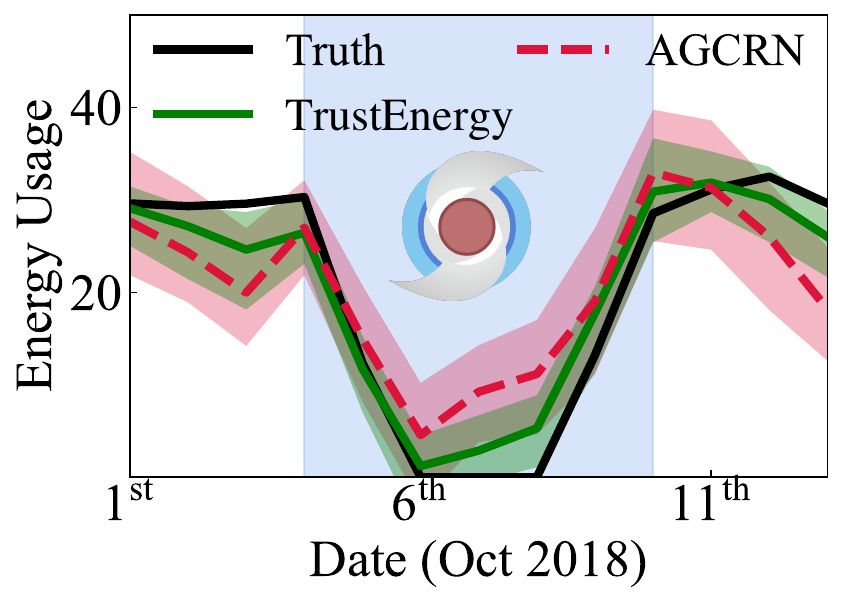}}
\subfigure[User 3 in Tornado]{
\includegraphics[width=0.48\linewidth]{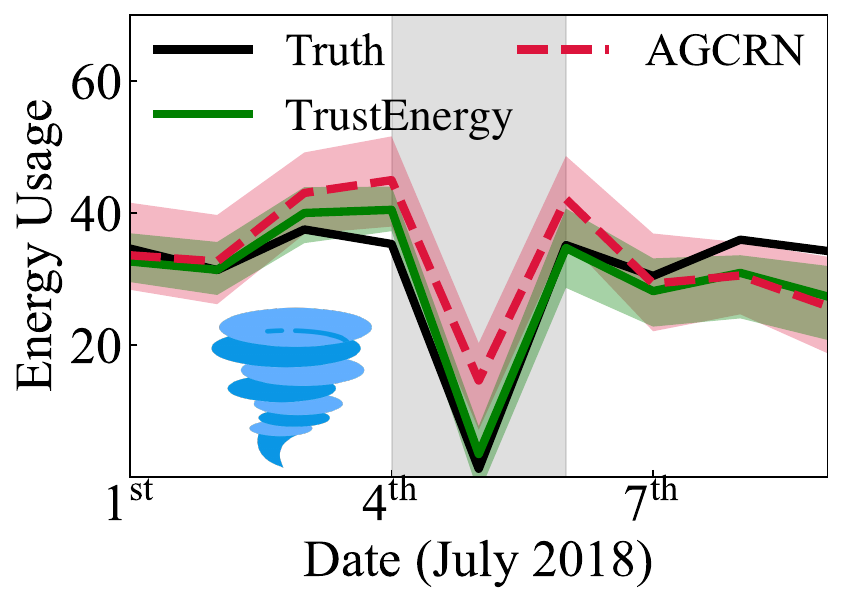}}
\subfigure[User 4 in Tornado]{
\includegraphics[width=0.48\linewidth]{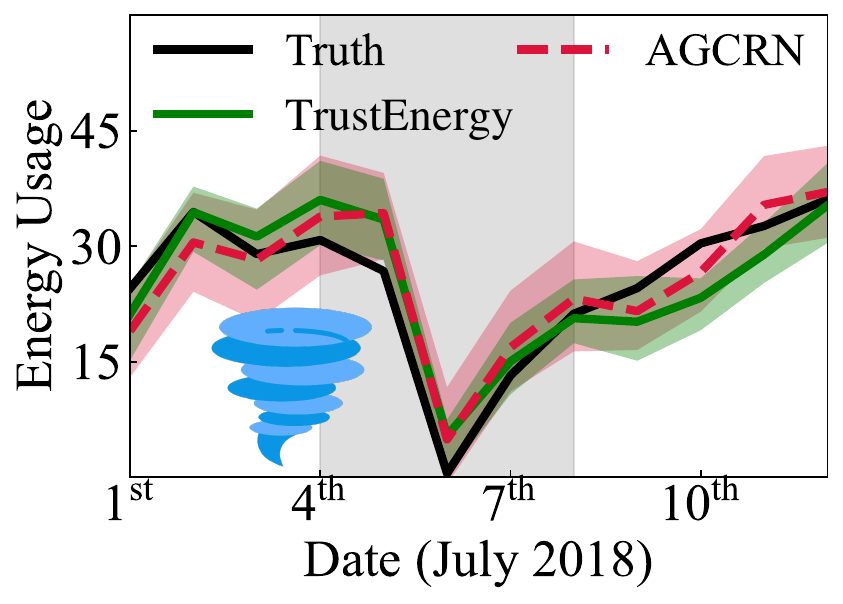}}
\subfigure[User 5 in Heat Wave]{
\includegraphics[width=0.48\linewidth]{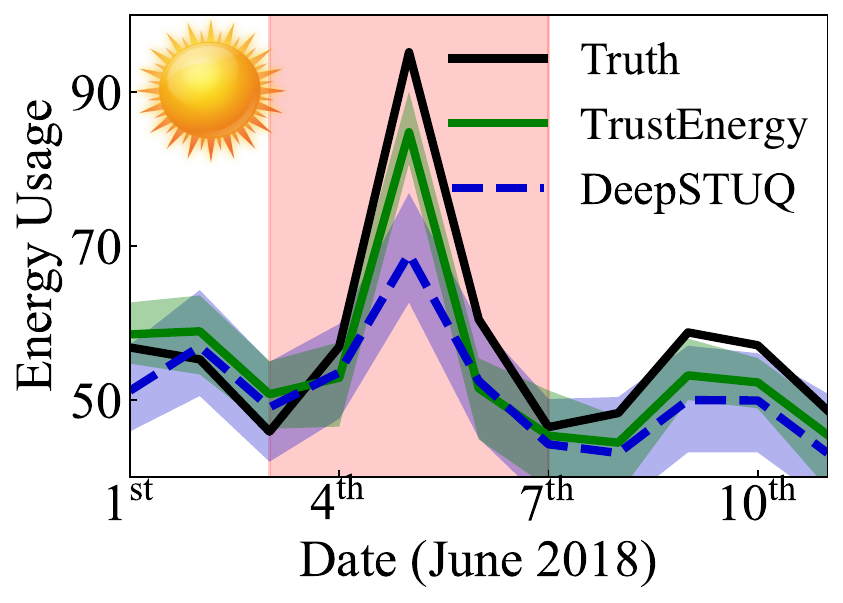}}
\subfigure[User 6 in Heat Wave]{
\includegraphics[width=0.48\linewidth]{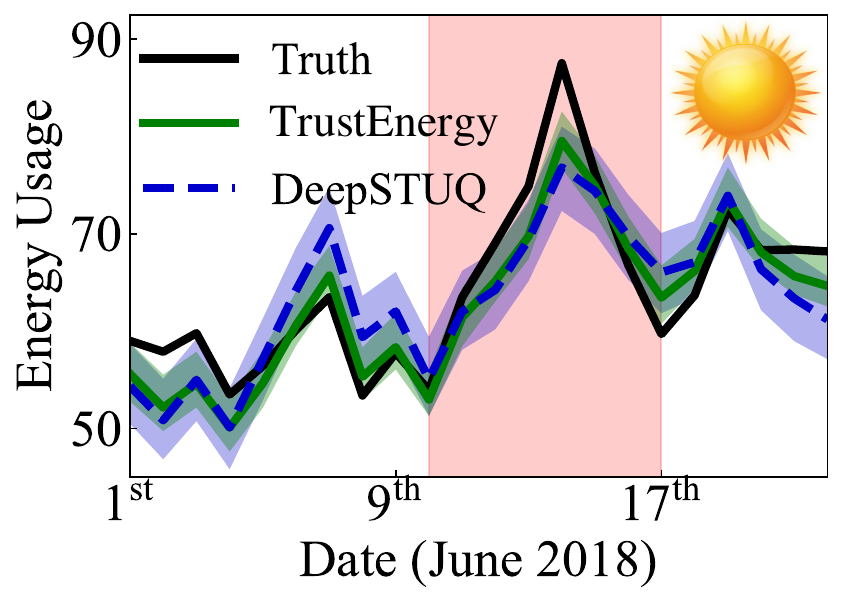}}
\caption{Prediction results under diverse extreme weather conditions.}
\label{hurricane}
\end{figure}

We utilize three representative extreme weather events: a major hurricane, a tornado, and an extreme heat wave in Florida, to show the effectiveness of our method.
As shown in Figure~\ref{hurricane}, \m maintained accurate predictions during the hurricane, with ground truth values mostly falling within the prediction intervals, while AGCRN failed to capture abrupt energy usage changes.
Similarly, under the heat wave scenario, where energy demand surged, \m outperformed DeepSTUQ with tighter intervals.
These results demonstrate \m's superior adaptability and robustness under extreme conditions.

\subsection{Ablation Study (RQ 3)}

We conduct an ablation study to evaluate the contribution of each key component in \m. Specifically, we compare \m with its following three variants:
\begin{itemize}
\item \textbf{w/o HSTR}: Replace hierarchical spatiotemporal representation module with only macro-level embeddings.
\item \textbf{w/o MASTGNN}: Remove the memory-augmented parameter update mechanism and use the conventional spatiotemporal learning module with DGCN and TCN.
\item \textbf{w/o SCQR}: Replace SCQR with traditional CQR, eliminating sequential dependency modeling.
\end{itemize}

\begin{table}[tb] \small
\setlength{\tabcolsep}{2pt}
\centering
\begin{tabular}{c|cccccc}
\toprule
 \multirow{2}{*}{Methods} & \multicolumn{5}{c}{Metrics} \\\cline{2-7}
		& MAE & RMSE& MAPE& MPIW & WINK &COV \\
\hline
 w/o HSTR& 49.35 & 47.35 & 1.48\% & 91.27 & 114.78 & \false \\
w/o MASTGNN& 47.54 & 57.88 & 1.33\% & 87.26 & 108.51 & \false \\
w/o SCQR& - & - & - & 87.05 & 104.70 & \false \\
\textbf{\m} & \textbf{45.72} & \textbf{55.65} & \textbf{1.28\%} & \textbf{85.72} & \textbf{102.31} & \true \\ 

\bottomrule
\end{tabular}
\caption{Ablation study of key design components.} 
\label{ablation}
\end{table}
From Table \ref{ablation}, we found that the model performance without the HSTR module significantly decreases (e.g., more than 7.3\% reduction in MAE on the Florida dataset), which indicates the importance of capturing comprehensive micro-level and macro-level patterns.
Similarly, by removing the memory-augmented parameter update mechanism, we found that the performance also decreased in all metrics (e.g., 3.8\% reduction in MAE).
In addition, we found that the uncertainty quantification performance decreased without the SCQR module since the prediction results cannot reach the target coverage.

\subsection{Generalizability to Other Areas (RQ 4)}

\begin{table}[tb]\small
\setlength{\tabcolsep}{1.5pt}
\centering 

\begin{tabular}{l|l|cccccc}
\toprule

\multirow{2}{*}{Dataset} & \multirow{2}{*}{Models} & \multicolumn{6}{c}{Metrics}\\ \cline{3-8}

& & MAE & RMSE & MAPE & MPIW & WINK & COV \\
\hline
\multirow{11}{*}{NY}
& STGCN     & 0.492 & 49.58 & 0.99\% & 3.328 & 3.331 & \false \\
& GWNET     & 0.199 & 8.364 & 0.60\% & 0.934  & 1.234 & \false \\
& ASTGCN    & 0.149 & \textbf{3.311} & 0.58\% & 0.558  & 0.851  & \false \\
& AGCRN     & \underline{0.147} & 3.361 & \underline{0.57\%} & \textbf{0.522}  & \underline{0.746}  & \false \\
& StemGNN   & 0.275 & 7.828 & 0.94\% & 0.888  & 1.437 & \true \\
& DSTAGNN   & 0.181 & 4.738 & 0.74\% & 0.774  & 1.045  & \false \\
& STZINB    & 0.149 & 3.825 & \textbf{0.54\%} & 0.588  & 0.781  & \true \\
&PDFormer & 0.243 & 8.352 & 0.86\% & 0.760 & 1.378 & \false \\
& DiffSTG   & 0.281 & 4.863 & 0.94\% & 1.108  & 1.751 & \true \\
& DeepSTUQ  & 0.310 & 5.132 & 0.98\% & 1.205  & 1.868  & \false \\
&PowerPM & 0.298 & 4.953 & 0.98\% & 0.854 & 1.361 & \false \\
&Chronos  &0.158& 4.365 & 0.59\% & 0.574 & 1.353& \true  \\
&Moment   &0.192& 5.174&  0.63\% & 0.623& 1.268& \false \\
& \textbf{\m} & \textbf{0.137} & \underline{3.321} & \underline{0.57\%} & \underline{0.540} & \textbf{0.727} & \true \\
\hline

\multirow{11}{*}{CA}
& STGCN     & 0.607 & 33.27 & 1.36\% & 3.500 & 3.776 & \true \\
& GWNET     & 0.353 & 5.565 & \underline{1.15\%} & 1.243 & \textbf{2.155} & \false \\
& ASTGCN    & 0.344 & \underline{3.142} & 1.28\% & 0.973 & 2.353 & \false \\
& AGCRN     & \underline{0.328}  & 3.256 & 1.24\% & \underline{0.854} & 2.303 & \false \\
& StemGNN   & 0.416 & 5.090 & 1.45\% & 1.575 & 2.475 & \true \\
& DSTAGNN   & \textbf{0.327}  & 4.014 & 1.23\% & \textbf{0.848} & 2.345 & \false \\
& STZINB    & 0.350  & 3.606 & 1.41\% & 0.905 & 2.471 & \false \\
&PDFormer & 0.381 & 4.128 & 1.39\% & 1.274 & 2.484 & \false \\
& DiffSTG   & 0.358 & 4.034 & 1.29\% & 1.101 & 2.254 & \true \\
& DeepSTUQ  & 0.384  & 4.152 & 1.36\% & 1.268 & \underline{2.225} & \false \\
&PowerPM & 0.337 & 4.128 & 1.28\% & 0.957 & 2.261 & \false \\
&Chronos     &0.348& 3.816 & 1.28\% & 0.944 & 2.235& \true  \\
&Moment      &0.352& 4.013&  1.29\% & 1.035& 2.274& \true \\
& \textbf{\m} & \textbf{0.327} & \textbf{3.012} & \textbf{1.14\%} & 0.896 & 2.325 & \true \\
\bottomrule
\end{tabular}
\caption{Comparison with 13 state-of-the-art baselines on two other datasets from New York and California.}\label{result_pub}
\end{table}

To verify the generalizability of our \m to different geographical areas, we also evaluate it on the NY and CA datasets. 
As shown in Table~\ref{result_pub}, \m consistently outperforms state-of-the-art baselines on most metrics, demonstrating its effectiveness and generalizability to broader areas. Also, the ablation study in Table~\ref{ablation_public} validates the utility of each component in our \m since the performance of the proposed framework will decrease without any key module.

\begin{table}[tb] \small
\setlength{\tabcolsep}{1.0pt}
\centering

\begin{tabular}{l|c|cccccc}
\toprule
\multirow{2}*{Dataset} & \multirow{2}{*}{Methods} & \multicolumn{5}{c}{Metrics} \\\cline{3-8}
		~ &  & MAE & RMSE& MAPE& MPIW & WINK &COV \\
\hline
\multirow{4}{*}{NY} & w/o HSTR& 0.431 & 7.583 & 1.26\% & 1.459 & 1.952 & \false \\
 & w/o MASTGNN& 0.358 & 5.752 & 1.15\% & 1.357 & 1.864 & \false \\
 & w/o SCQR& - & - & - & 0.875 & 1.054 & \true \\
 & \textbf{\m} & \textbf{0.137} & \textbf{3.321} & \textbf{0.57\%} & \textbf{0.540} & \textbf{0.727} & \true \\ 
\hline

\multirow{4}{*}{CA} & w/o HSTR& 0.471 & 7.713 & 1.43\% & 1.368 & 2.675 & \false \\
 & w/o MASTGNN& 0.405 & 5.257 & 1.39\% & 1.257 & 2.568 & \false \\
 & w/o SCQR& -& - & - & 1.125 & 2.463 & \true \\
 & \textbf{\m} & \textbf{0.327} & \textbf{3.012} & \textbf{1.14\%} & \textbf{0.896} & \textbf{2.325} & \true \\ 
 
\bottomrule
\end{tabular}

\caption{Ablation study on NY and CA datasets.} \label{ablation_public}
\end{table}

\section{Related Work}\label{literature review}

\subsection{Energy Usage Prediction}
As an important real-world problem, energy usage prediction has attracted much interest from both academia and industry. Traditional machine learning methods have long been employed, such as linear regression~\cite{hong2011naive}, support vector regression~\cite{sapankevych2009time}, decision tree-based models~\cite{lloyd2014gefcom2012}, and random forest regression~\cite{wu2015power}.
However, these methods often struggle to capture the complex, nonlinear relationships inherent in energy usage data. 
In recent years, more advanced deep learning methods~\cite{luo2024stacking, balachander2024innovative, buratto2024seq2seq} have been leveraged to predict energy usage by leveraging sophisticated deep learning architectures, but most of them focus on aggregated time series prediction while ignoring rich user-level and spatial information, which undermines their capabilities for making accurate fine-grained predictions. 
In addition, energy usage prediction reliability will directly impact real-world decision-making, such as resource allocation or restoration under extreme weather conditions. However, this topic has not been fully explored by existing work.

\subsection{Spatiotemporal Uncertainty Quantification}
Recent efforts on spatiotemporal uncertainty quantification provide a good opportunity to quantify energy usage prediction reliability. Wen \textit{et al.}~\cite{wen2023diffstg} extend denoising diffusion probabilistic models to spatiotemporal graphs via DiffSTG for capturing intrinsic uncertainties. 
Kexin \textit{et al.}~\cite{huang2024uncertainty} adapt conformal prediction to graph-based models to provide valid uncertainty guarantees. Zhuang \textit{et al.}~\cite{stzinb-gnn} propose STZINB, a zero-inflated negative binomial GNN designed for modeling uncertainty in sparse urban data. Qian \textit{et al.}~\cite{uqunifi} present DeepSTUQ, which leverages dual neural subnetworks to estimate uncertainty in traffic prediction tasks.

Nevertheless, most existing methods~\cite{xu2023quantile, faustine2022fpseq2q} for spatiotemporal uncertainty quantification either impose restrictive assumptions on spatial and temporal dependencies or fail to jointly model the interplay between them, which may cause the prediction intervals to be poorly calibrated, and they cannot dynamically adjust uncertainty bounds over time to adapt to the distribution shifts caused by factors like extreme weather events.

\section{Conclusion}\label{conclusion}
In this work, we design a unified framework called \m to improve both the accuracy and reliability of user-level energy usage prediction. 
There are two key novel designs in \m, i.e., (i) a hierarchical spatiotemporal representation module to efficiently capture both
macro and micro patterns with a new MASTGNN, and (ii) an innovative distribution-agnostic uncertainty quantification method called SCQR to dynamically adjust uncertainty bounds to ensure valid prediction intervals over time.
By collaborating with an electricity provider in Florida, we extensively evaluate \m by comparing it against 13 state-of-the-art baselines across six metrics. The results demonstrate \m effectively outperforms baselines and improves prediction accuracy by around 5.4\% and prediction reliability by around 5.7\%. Notably, \m proves to be particularly effective under extreme weather conditions such as hurricanes, highlighting its potential for real-world societal impact. We also validate its generalizability using real-world datasets from New York and California, which also indicate the effectiveness of \m.


\section*{Acknowledgments}
We sincerely thank all anonymous reviewers for their insightful comments and valuable suggestions. This work is partially supported by the FSU Sustainability \& Climate Solutions Grant Program, FSU/AWS Computer Support Seed Fund, and FSU Startup Fund. Shenhao Wang acknowledges the support from UFL's 2023 Research Opportunity Seed Fund (ROSF2023). 
\bibliography{reference}



\end{document}